\documentclass{article}
\usepackage{ijcai26}

\usepackage{times}
\usepackage{soul}
\usepackage{url}
\usepackage[hidelinks]{hyperref}
\usepackage[utf8]{inputenc}
\usepackage[small]{caption}
\usepackage{graphicx}
\usepackage{amsmath}
\usepackage{amsthm}
\usepackage{booktabs}
\usepackage{algorithm}
\usepackage{algorithmic}
\usepackage[switch]{lineno}
\usepackage{multirow}
\usepackage{makecell}
\usepackage{subcaption}
\usepackage{amssymb}
\usepackage{tabularx}
\usepackage{array}
\usepackage[table]{xcolor}
\definecolor{lightgray}{gray}{0.9}

\title{PIS: A Physics-Informed System for Accurate State Partitioning of $A\beta42$ Protein Trajectories}

\author{
Qianfeng Yu$^1$\and
Ningkang Peng$^1$\and
Yanhui Gu$^{1}$\thanks{Corresponding authors.}
\\
\affiliations
$^1$School of Computer and Electronic Information, Nanjing Normal University\\
\emails
\{qfyu, nkpeng\}@nnu.edu.cn, gu@njnu.edu.cn
}

\begin{document}

\maketitle

\begin{abstract}
Understanding the conformational evolution of $\beta$-amyloid ($A\beta$), particularly the $A\beta_{42}$ isoform, is fundamental to elucidating the pathogenic mechanisms underlying Alzheimer's disease. However, existing end-to-end deep learning models often struggle to capture subtle state transitions in protein trajectories due to a lack of explicit physical constraints. In this work, we introduce PIS, a Physics-Informed System designed for robust metastable state partitioning. By integrating pre-computed physical priors, such as the radius of gyration and solvent-accessible surface area, into the extraction of topological features, our model achieves superior performance on the $A\beta_{42}$ dataset. Furthermore, PIS provides an interactive platform that features dynamic monitoring of physical characteristics and multi-dimensional result validation. This system offers biological researchers a powerful set of analytical tools with physically grounded interpretability. A demonstration video of PIS is available on \url{https://youtu.be/AJHGzUtRCg0}.

\end{abstract}
\section{Introduction}
Alzheimer's disease is a leading cause of global mortality, characterized pathologically by the formation of neurotoxic $\beta$-amyloid ($A\beta$) aggregates \cite{nasica2015amyloid,meng2018highly}, particularly the $A\beta_{42}$ isoform, within the brain. Evidence suggests that this aggregation involves a series of non-linearly coupled microscopic steps, making the investigation of $A\beta_{42}$ monomer evolution essential. However, $A\beta_{42}$ is a prototypical intrinsically disordered protein (IDP) that exhibits extreme conformational heterogeneity. Since experimental methods struggle to capture the transient evolution of $A\beta_{42}$ at microsecond timescales, molecular dynamics (MD) simulations have become the primary tool for exploring the $A\beta_{42}$ conformational ensemble.

The prevailing paradigm for establishing kinetic models of $A\beta_{42}$ combines Variational Approach to Conformational Dynamics (VAC) or Variational Markov Processes (VAMP) with deep neural networks, enabling end-to-end feature learning from raw trajectories to metastable state partitioning. Pioneering work in this domain includes VAMPNet \cite{mardt2018vampnets}. Subsequently, RevDMSM \cite{lohr2021kinetic} imposed reversibility and non-negativity constraints on the transition matrix \cite{husic2018markov}, introducing the VAMP-E loss function to optimize model performance. GraphVAMPNet \cite{ghorbani2022graphvampnet} further enhanced dynamical resolution by employing graph neural networks, such as SchNet \cite{schutt2017schnet}, to capture spatial topology alongside attention mechanisms. More recently, RevGraphVAMP \cite{huang2024revgraphvamp} fused graph convolutional networks with constrained optimization, utilizing its GASchNet component to extract structural features. This approach substantially improved predictive accuracy for non-equilibrium sampled trajectories and the distinguishability of metastable states. Despite these advancements in incorporating physical constraints, existing models still encounter ambiguity in critical state partitioning when processing $A\beta_{42}$ trajectories.

\begin{figure}[t!]
    \centering \includegraphics[width=0.96\columnwidth]{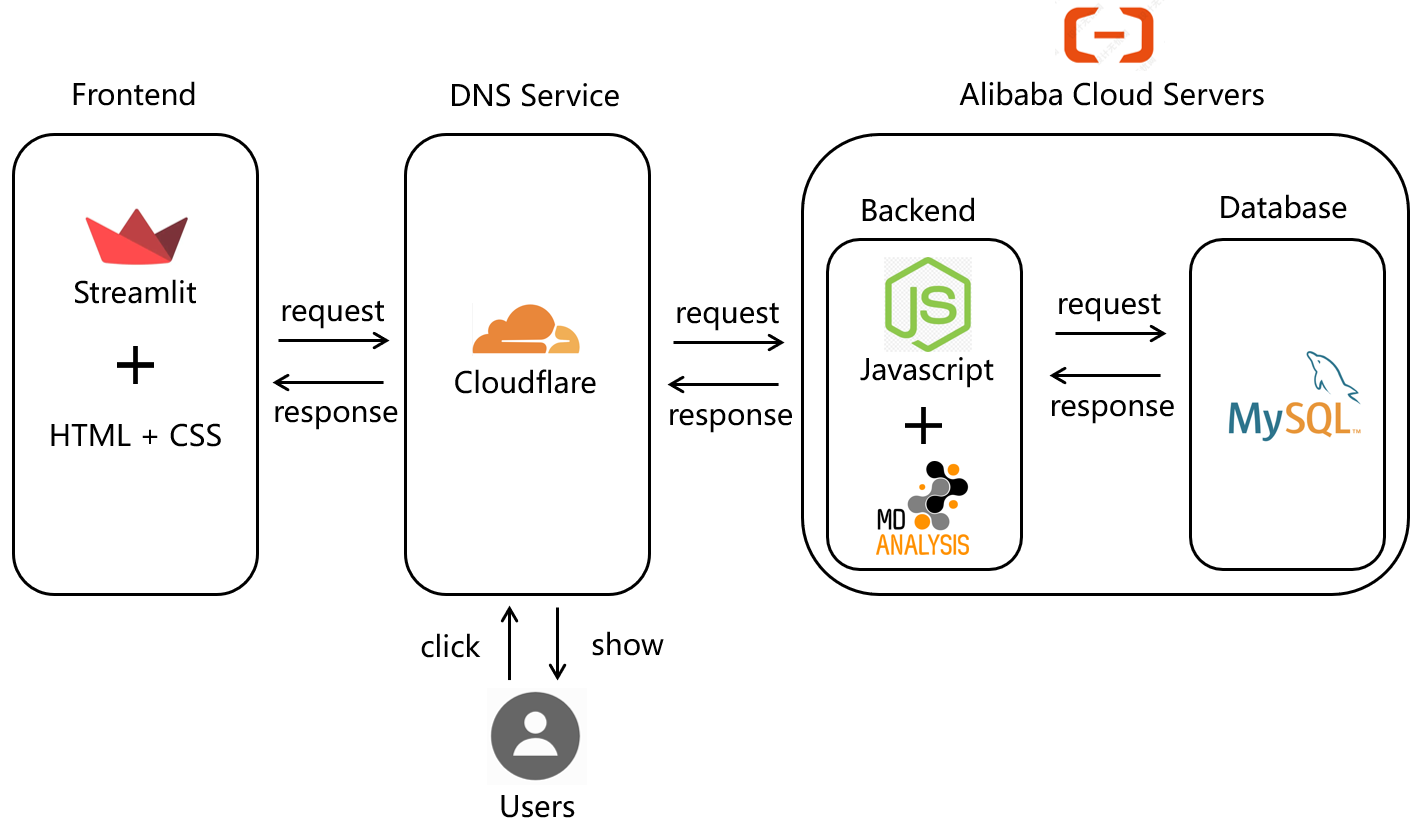} 
    \caption{The architecture of PIS.}
    \label{fig:architecture}
\end{figure}

To address these challenges, we propose PIS, a Physics-Informed System for the accurate state partitioning of $A\beta_{42}$ protein trajectories. By integrating pre-computed physical priors—such as the radius of gyration ($R_g$) and solvent-accessible surface area (SASA)—during topological feature extraction, PIS enhances performance on VAMP-2 and VAMP-E metrics \cite{wu2020variational}, thereby compensating for the lack of physical intuition in purely data-driven models. Concurrently, PIS offers an interactive platform that provides dynamic monitoring of physical features and multi-dimensional result validation, furnishing biological researchers with an analytical tool set characterized by interpretable physical metrics. Figure~\ref{fig:architecture} illustrates the architecture of PIS, detailing the primary components of its frontend and backend systems and their interrelationships.

\section{Methodology}
\begin{figure}[t!]
    \centering \includegraphics[width=0.96\columnwidth]{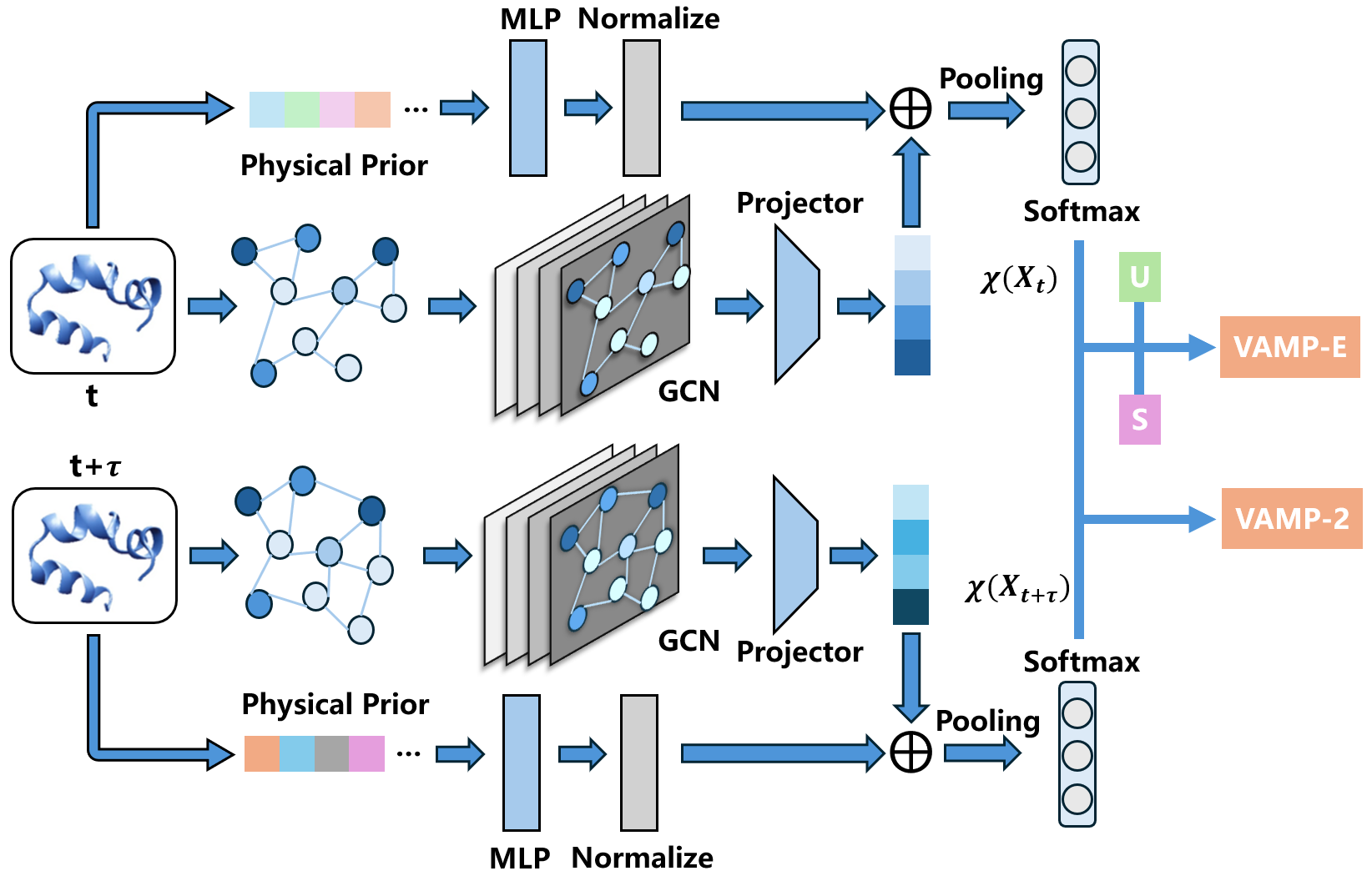} 
    \caption{The overall framework of the PIS model.}
    \label{fig:framework}
\end{figure}

The framework of PIS is illustrated in Figure~\ref{fig:framework}. We present a concise overview of the methodology and detail each component in the following sections.

\subsection{Data Source}
We utilize a publicly available all-atom MD dataset of $A\beta_{42}$ \cite{lohr2021kinetic} to evaluate the performance of our system. Simulations were conducted using GROMACS 2018.1 with the CHARMM22 force field, incorporating approximately 11,751 TIP3P water molecules. With an integration time step of 2 fs and a storage interval of 250 ps, the dataset encompasses an aggregate simulation time of 315 $\mu$s, comprising 5,119 independent trajectories and 1,259,172 sampled frames. In the feature extraction module, we model 42 core residues by identifying the 10 nearest neighbors for each residue based on minimal heavy-atom distances to compute edge embeddings. The massive scale of this dataset serves as a rigorous benchmark for assessing the robustness of our visualization system in handling high-throughput biological simulation data.

\subsection{Dual-Track Fusion}
The system initially models the protein conformation as an attributed graph $G = (V, E)$, where nodes $V$ represent the 42 residues and edges $E$ capture their spatial interdependencies. To precisely characterize microscopic interactions, edge representations are derived by applying a Gaussian diffusion function to the heavy-atom distances $d_{ij}$. Information exchange is subsequently performed via the GASchNet component, which iteratively updates node states through continuous filter convolutions. Attention weights are incorporated during this process to capture critical interaction regions. Finally, a global pooling operation is applied to extract the microscopic topological embedding vector $\mathbf{v}_G$.

To mitigate the ambiguity inherent in purely data-driven models during transition state classification, we integrate a global physical feature stream $\mathbf{p}$, encompassing key indicators such as the radius of gyration ($R_g$) and solvent-accessible surface area (SASA). Inspired by dual-track feedback architectures, we implement a gated adaptive addition mechanism for deep feature fusion. Initially, raw physical indicators are mapped via a linear layer with ReLU activation, followed by normalization to yield the preprocessed physical vector $\mathbf{p}'$:
\begin{equation}
    \mathbf{p}' = \text{Normalize}(\sigma(\mathbf{W}_1 \mathbf{p} + \mathbf{b}_1)).
\end{equation}

A compact self-gating module then dynamically computes a fusion weight $\alpha \in (0, 1)$ based on the contextual information from the topological representation $\mathbf{v}_G$ and the physical prior $\mathbf{p}'$:
\begin{equation}
    \alpha = \sigma(\mathbf{W}_g [\mathbf{v}_G \parallel \mathbf{p}'] + b_g).
\end{equation}

This weight facilitates a weighted summation, enabling the injection of physical signals into the topological space while maintaining a concise feature representation:
\begin{equation}
    \mathbf{h}_{fuse} = \alpha \cdot \mathbf{v}_G + (1 - \alpha) \cdot \mathbf{p}'.
\end{equation}

Finally, the fused composite signal $\mathbf{h}_{fuse}$ is aggregated through an attention pooling mechanism to generate the final metastable state embedding vector $\mathbf{z}$.
Subsequently, a Softmax layer maps the representation into a classification probability distribution $\chi(x)$. The output value $\chi_{i}(x)$ denotes the probability that the protein conformation belongs to substate $i$ at time $t$, subject to the normalization constraint $\sum \chi_{i}(x) = 1$. For conformational analysis, each structure is categorized into the substate exhibiting the highest output value.

\subsection{Optimization Objectives}
For model optimization, a multi-stage training strategy is implemented leveraging both VAMP-2 and VAMP-E loss functions \cite{wu2020variational}. In the initial phase, the VAMP-2 score is employed for the unsupervised pre-training of the transformation function $\chi(x)$. The system then transitions to a training phase governed by VAMP-E as the primary loss function. The integration of VAMP-E facilitates the use of trainable $U$ and $S$ constraint modules, which enforce physical reversibility and statistical stationarity on the transition matrix $K$. As shown in Table~\ref{tab:ab42_scores}, we compared PIS against multiple benchmark models, including traditional VAMPnets \cite{mardt2018vampnets,lohr2021kinetic}, GraphVAMPNet's graph convolutional network (GCN-VAMP) \cite{ghorbani2022graphvampnet}, and the state-of-the-art RevGraphVAMP \cite{huang2024revgraphvamp}. Experimental results demonstrate the outstanding performance of our system when processing the highly challenging $A\beta_{42}$ disordered peptide dataset.

\begin{figure*}[t!]
    \centering \includegraphics[width=0.96\textwidth]{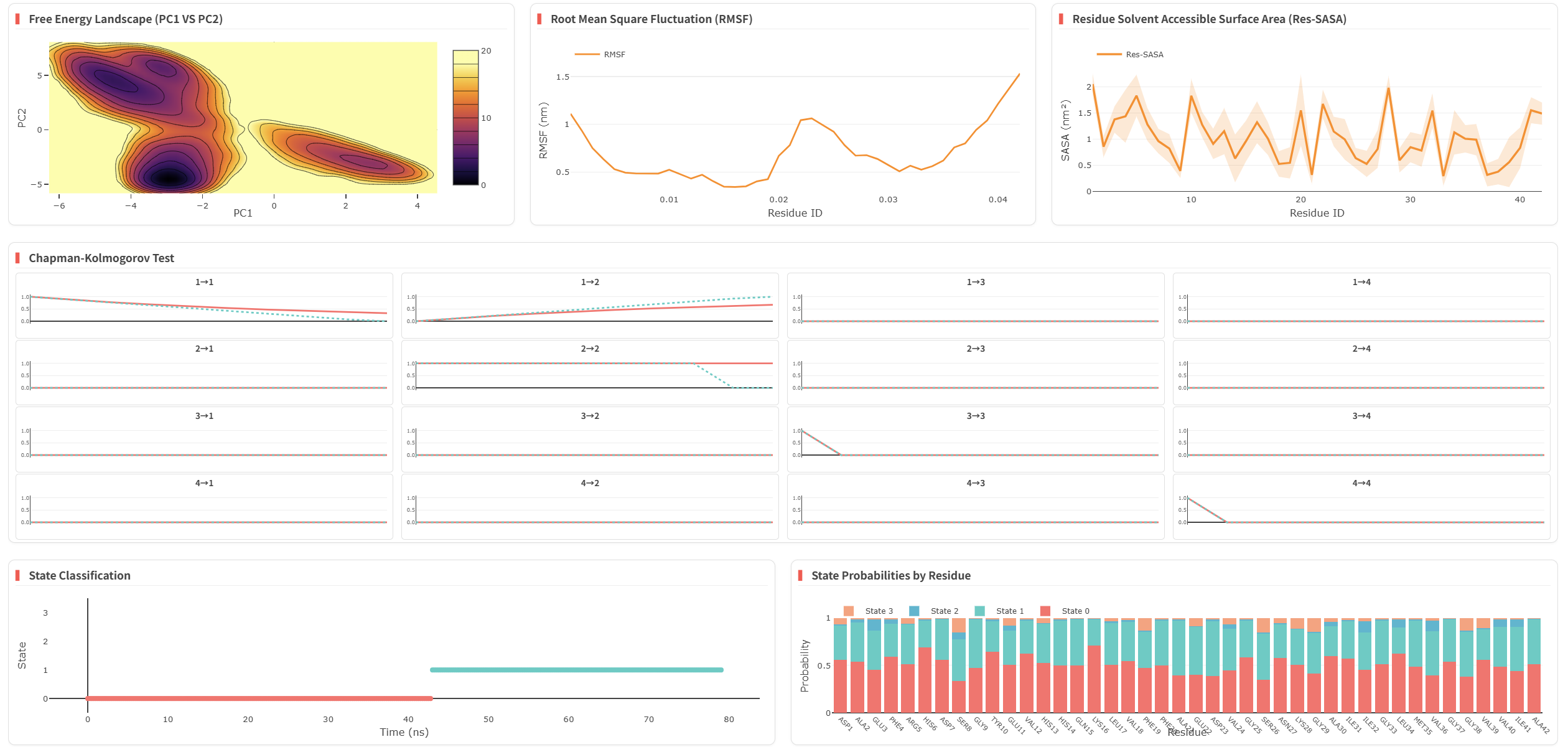} 
    \caption{User interface for multi-dimensional result validation and kinetic analysis.}
    \label{fig:multi}
\end{figure*}

\begin{table}[h]
    \centering
    \begin{tabular}{ccc}
    \toprule
    \textbf{Model} & \multicolumn{2}{c}{\textbf{$A\beta_{42}$}} \\
    \cmidrule{2-3}
    & VAMP-2 & VAMP-E \\ \midrule
    VAMPnets & $3.98 \pm 0.008$ & $3.93 \pm 0.050$ \\
    GCN-VAMP & $3.99 \pm 0.008$ & $3.97 \pm 0.060$ \\
    MAGNN-VAMP & $3.97 \pm 0.001$ & $3.95 \pm 0.002$ \\
    RevGraphVAMP & $\mathbf{3.99 \pm 0.002}$ & $3.98 \pm 0.003$ \\
    PIS & $\mathbf{3.99 \pm 0.003}$ & $\mathbf{3.99 \pm 0.003}$ \\\bottomrule
    \end{tabular}
    \caption{Comparison of VAMP scores for $A\beta$42.}
    \label{tab:ab42_scores}
\end{table}


\begin{figure}[t!]
    \centering \includegraphics[width=0.96\columnwidth]{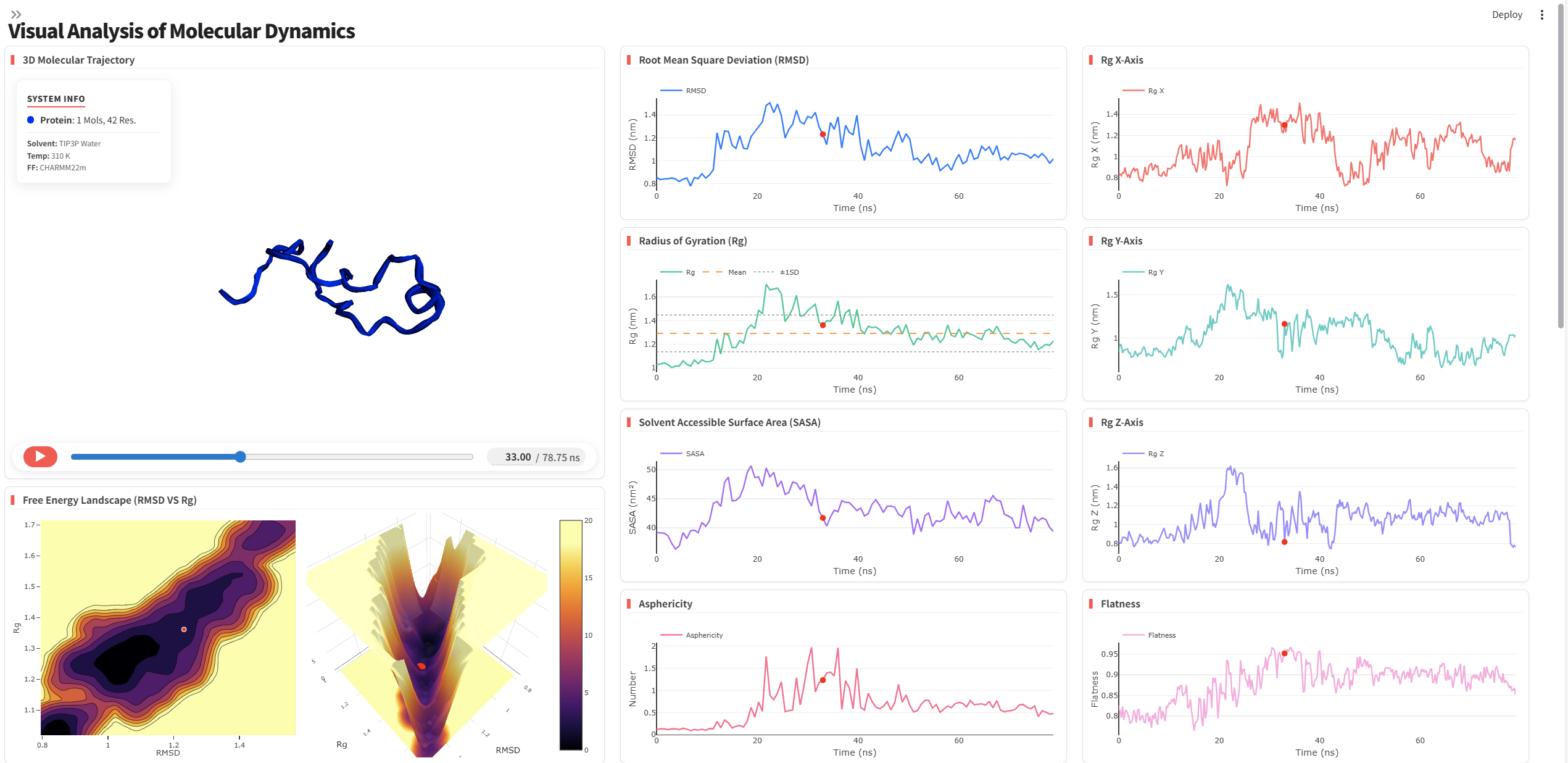} 
    \caption{Interface for the dynamic monitoring of conformational evolution and physical metrics.}
    \label{fig:dpm}
\end{figure}

\section{Demonstration}
To demonstrate the practical utility of PIS in analyzing $A\beta_{42}$ kinetic trajectories, we developed an integrated web-based interactive platform. This platform is designed to provide researchers with intuitive tools spanning from raw trajectory analysis to the validation of kinetic mechanisms.

\subsection{Dynamic Physics Monitoring}
To ensure a seamless user experience when processing the massive $A\beta_{42}$ dataset of over 1.2 million frames, PIS leverages a highly optimized WebGL rendering engine. This architecture supports low-latency, 60 FPS interactive playback of all-atom trajectories, as illustrated in Figure~\ref{fig:dpm}. Unlike static tools, PIS achieves sub-millisecond synchronization between 3D conformational transitions and dynamic physical metric fluctuations, enabling researchers to capture critical transition states via instantaneous visual feedback. Simultaneously, PIS employs real-time red markers to track values on physical metric curves, including the radius of gyration ($R_g$) and solvent-accessible surface area (SASA), corresponding to the current frame. Monitoring these fluctuations allows researchers to pinpoint critical physical transition points during protein conformational shifts, which is essential for understanding complex folding pathways.

\subsection{Multi-Dimensional Result Verification}
Building upon real-time monitoring, the platform offers in-depth statistical and physical validation modules to resolve complex $A\beta_{42}$ kinetic features from multiple perspectives, as illustrated in Figure~\ref{fig:multi}. For the free energy landscape, the system extracts the top two principal components ($PC_1, PC_2$) from the feature embedding space as reaction coordinates to dynamically generate surfaces, intuitively demonstrating the distinct separation of metastable basins after incorporating physical priors. Regarding mechanistic insight at the residue level, individual residue stability is monitored via Root Mean Square Fluctuation (RMSF) and residue-level solvent-accessible surface area (Res-SASA), while quantifying amino acid contribution probabilities identifies correlations between specific residues and conformational states. For the Chapman-Kolmogorov test, the platform plots predicted versus observed curves by averaging results across varying lag times, ensuring physical consistency and extrapolative accuracy of the transition matrix $K(\tau)$ at microsecond timescales. Finally, state classification allows researchers to instantaneously identify the kinetic ensemble of the current conformation while observing all-atom trajectories, enhancing the efficiency of complex trajectory analysis and providing deeper insights into metastable state transitions.



\bibliographystyle{named}
\bibliography{ijcai26}

\end{document}